\documentclass[10pt,twocolumn,letterpaper]{article}

\usepackage{cvpr}              
\usepackage[symbol]{footmisc}
\usepackage{lipsum}
\usepackage{multirow}

\usepackage{xspace}
\newcommand*{\sysname}{Focal Split\xspace}

\usepackage{pifont}
\definecolor{brightmaroon}{rgb}{0.76, 0.13, 0.28}
\newcommand{\xmark}{\textcolor{brightmaroon}{\ding{55}}}
\newcommand{\cmark}{\textcolor{green}{\ding{52}}}

\definecolor{cvprblue}{rgb}{0.21,0.49,0.74}
\usepackage[pagebackref,breaklinks,colorlinks,allcolors=cvprblue]{hyperref}
\usepackage{algorithm}
\usepackage{algpseudocode}
\usepackage{array}
\usepackage{cellspace}

\def\paperID{16668} 
\def\confName{CVPR}
\def\confYear{2025}

\title{Focal Split: Untethered Snapshot Depth from Differential Defocus}

\author{Junjie Luo$^{*,1}$, John Mamish$^{*,2}$, Alan Fu$^{*,1}$, Thomas Concannon$^{1}$, \\
Josiah Hester$^{2}$, Emma Alexander$^{3, \dagger}$, and Qi Guo$^{1, \ddagger}$\\
$^{1}$Elmore Family School of Electrical and Computer Engineering, Purdue University\\
$^{2}$College of Computing, Georgia Institute of Technology\\
$^{3}$McCormick School of Engineering, Northwestern University\\
 $^*$ Equal contributions, 
{\tt\small $^\dagger$ealexander@northwestern.edu, $^\ddagger$qiguo@purdue.edu}
}

\begin{document}

\newcommand{\qg}[1]{{\color{red} #1}}
\twocolumn[{%
\renewcommand\twocolumn[1][]{#1}%
\maketitle
\begin{center}
    \centering
    \captionsetup{type=figure}
    \vspace{-0.2in}
    \includegraphics[width=1.00\textwidth]{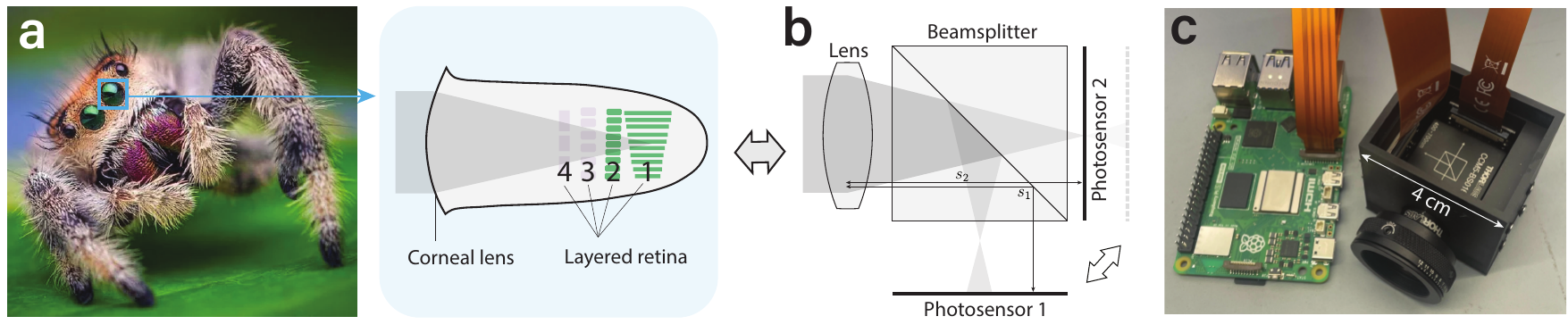}
    \captionof{figure}{Overview. (a) The principal eyes of jumping spiders comprise layered retinae, allowing the same scene to be imaged simultaneously at slightly different distances from the lens. This enables them to see two differentially defocused images of a target, from which depth can be estimated efficiently~\cite{nagata2012depth}. 
    (b) \sysname's novel optomechanical setup leverages a beamsplitter and two photosensors placed at different sensor distances to the lens to mimic the jumping spider's eye structures. 
    (c) Our handheld, untethered \sysname prototype can generate real-time sparse depth maps from battery-powered on-board computing.}
    \label{fig:teaser}
\end{center}%
}]

\begin{abstract}

We introduce Focal Split, a handheld, snapshot depth camera with fully onboard power and computing based on depth-from-differential-defocus (DfDD). 
Focal Split is passive, avoiding power consumption of light sources.
Its achromatic optical system simultaneously forms two differentially defocused images of the scene, which can be independently captured using two photosensors in a snapshot. The data processing is based on the DfDD theory, which efficiently computes a depth and a confidence value for each pixel with only 500 floating point operations (FLOPs) per pixel from the camera measurements. We demonstrate a Focal Split prototype, which comprises a handheld custom camera system connected to a Raspberry Pi 5 for real-time data processing. The system consumes 4.9 W and is powered on a 5 V, 10,000 mAh battery. The prototype can measure objects with distances from 0.4 m to 1.2 m, outputting 480$\times$360 sparse depth maps at 2.1 frames per second (FPS) using unoptimized Python scripts. Focal Split is DIY friendly. A comprehensive guide to building your own Focal Split depth camera, code, and additional data can be found at \url{https://focal-split.qiguo.org}. 





\end{abstract}    
\section{Introduction}



\textit{Depth from differential defocus (DfDD)} is a family of physically rigorous depth-sensing methods that generate a depth map from a series of \textit{differentially defocused} images with extremely efficient computations~\cite{alexander2019theory}. Its development was partially inspired by the optical functionalities of jumping spider's eyes~\cite{land1969structure,nagata2012depth}. Here, we address two major shortcomings in this family. First, existing DfDD cameras either require sequential image captures and assume the scene to be static~\cite{guo2017focal,luo2024depth}, additional computation to handle motion between frames~\cite{alexander2016focal,alexander2018focal}, or narrowed illumination bandwidth in conjunction with a multifunctional metasurface~\cite{guo2019compact}. In contrast to the snapshot multi-focus capture of the spider's layered retina (Fig.~\ref{fig:teaser}a), this limits their accuracy or light efficiency in the face of moving scenes under normal illumination. Second, despite the promise of DfDD algorithms for low-power applications, these prototypes have all been tethered to wall sockets for power~\cite{guo2017focal} and to laptop computers for depth map calculation~\cite{alexander2016focal, guo2017focal, guo2019compact, luo2024depth}.


We present \textit{Focal Split}, the first snapshot, untethered DfDD camera. As shown in Fig.~\ref{fig:teaser}c, Focal Split utilizes a novel optical design that functionally mimics jumping spiders' eye structure. It splits the incident light focused by a lens through a beamsplitter to form two images with different sensor distances $s_1$ and $s_2$. By synchronizing the two photosensors, the hardware captures a pair of images, $I_1$ and $I_2$, with varied sensor distances simultaneously, effectively resembling a pair of layered semi-transparent retinae. Compared to previous DfDD sensors that sequentially capture the differently defocused images by varying the focal length of the lens~\cite{guo2017focal}, aperture diameter~\cite{schechner2000depth}, aperture code~\cite{zhou2011coded}, or camera positions~\cite{alexander2018focal}, Focal Split achieves the critical advantage of capturing $I_1$ and $I_2$ in a snapshot. This avoids misalignment between the images when calculating the depth map, which, as we will show, seriously contaminates the quality of the depth maps.

We also derive a new physically rigorous depth estimation algorithm specialized for the optical setup of this paper. It shows that the depth map of the scene can be calculated by a simple pixel-wise expression with the image derivatives:
\begin{align}
    Z = \frac{a}{b + I_s / \nabla^2 I},
    \label{eq:dfdd-intro}
\end{align}
where $a$ and $b$ are constants determined by the optics and $I_s$ and $\nabla^2 I$ are the image derivatives that can be estimated from the pair of differently defocused images, $I_1$ and $I_2$, after aligning their magnification. The proposed algorithm is a novel instance that belongs to the family of depth from differential defocus~\cite{alexander2019theory}. 
In this paper, we perform comprehensive theoretical and simulation analyses on the proposed algorithm's sensitivity to texture frequency, noise, etc. 

Our algorithm can only produce a partially dense depth map from the input images. This is because Eq.~\ref{eq:dfdd-intro} degenerates at textureless regions of the image ($I_s$ and $\nabla^2 I$ becomes zero.) In fact, all passive-ranging methods, including stereo, fundamentally fail at textureless regions. Previous algorithms typically perform passive ranging and implicit depth map densification in a single model. These methods cost relatively high computation but produce high-quality, dense depth maps. Compared to them, our algorithm provides an alternative option to generate a partially dense depth map with a much lower computation, leaving the densification to downstream tasks, which is suitable for low-power, autonomous platforms, such as micro-robots, autonomous underwater vehicles, AR glasses, etc. Using Eq.~\ref{eq:dfdd-intro}, Focal Split only costs 500 floating point operations (FLOPs) per pixel to generate a partially dense depth map.




Focal Split is the first untethered DfDD camera. As shown in Fig.~\ref{fig:teaser}c, this handheld system comprises a custom housing of the optics and photosensors and a Raspberry Pi 5 that performs real-time onboard depth estimation. The housing is 4 cm $\times$ 5 cm $\times$ 6 cm. The system can output depth maps at 480$\times$360 resolution at 2.1 FPS using un-optimized Python scripts with the power consumption of only 4.9 W. 
The contributions of this work can be summarized as follows:

\begin{enumerate}
    \item \textbf{A new optical design} that functionally mimics the principle eye of jumping spiders. It enables the simultaneous capture of a pair of differentially-defocused images. 
    \item  \textbf{A new DfDD algorithm} with a verified advantage in robustness for our optical design. 
    \item \textbf{A new low-power, untethered working prototype for snapshot depth sensing } with a power budget $<10$ W and a significant error reduction compared to 
    sequential measurement DfDD for dynamic scenes.
\end{enumerate}

\section{Related Work}


Techniques for depth imaging can broadly be divided into two categories: active and passive. Generally speaking, techniques from each of these categories have somewhat similar characteristics; active imaging systems typically have worse size, weight, power, and cost (SWaP-C) but better accuracy, while passive imaging systems being complementary with better SWaP-C and worse accuracy~\cite{dong2022towards}. Below, we will give a discussion of power consumption and portability for depth imaging systems using both active and passive techniques.

\subsection{Active techniques}
``Active imaging" is a descriptor for any system which must project light into a scene in order to image it. Because of their accuracy, active imaging systems are considered to be the gold standard for depth imaging.
However, due to the power needed to illuminate scenes, active imagers are not appropriate for low-power, portable applications~\cite{dong2022towards}.

\noindent\textbf{LiDAR.} LiDAR systems are active imaging systems which work by sending laser pulses into a scene and recording their round-trip time-of-flight with high precision to calculate distance. However, high-powered lasers are required for LiDAR systems to operate, limiting their portability~\cite{dong2022towards}. Furthermore, LiDAR systems require their light source to be scanned across a scene, meaning that bulky and power-consuming opto-mechanical setups must be used ~\cite{raj2020survey}. Some recent works improve the power consumption of LiDAR systems by introducing novel scanning mechanisms~\cite{pittaluga2020towards} and deep-learning based methods for depth completion on sparse maps~\cite{tasneem2020adaptive,bergman2020deep}, but these methods depend on novel opto-mechanical components and large DNNs, limiting their adoption and restricting their use in edge systems.


\noindent\textbf{Structured Light.} Structured light 3D scanners operate by illuminating a scene with a specifically chosen pattern and imaging the resulting scene. Because the projected light pattern is known, the measured points can be used to calculate depth~\cite{intel:realsense,geng2011structured}. Structured light systems require less power than LiDAR systems~\cite{intel:realsense} because they do not require nanosecond-level timing precision, but they are less robust and require sophisticated image processing pipelines to calculate depth images~\cite{geng2011structured}.

\subsection{Passive techniques}
Unlike active imaging systems, passive depth imagers do not emit any light. This removes the most energy intensive component of active systems, but passive depth imagers often require more sophisticated image processing pipelines and are less robust~\cite{kazerouni2022survey}.

\noindent\textbf{Monocular Structure from Motion (SfM).} Structure from motion (SfM) refers to the recovery of 3D structure from a sequence of monocular images~\cite{puglia2017real,ozyecsil2017survey}. Although SfM techniques are appealing because their monocular image sensors are compact and cheap~\cite{ozyecsil2017survey}, viable SfM methods are computationally intensive and require significant memory, restricting their adoption in portable, low-power, real-time systems~\cite{mouragnon2009generic,schonberger2016structure,jiang2020efficient}.


\noindent\textbf{Stereo vision.} Stereo vision is one of the most well-researched methods for producing depth images from conventional image sensors. By positioning 2 cameras a fixed distance apart, depth can be recovered by looking at the differences in matching scene features' positions between the images. 
While conceptually simple, this problem is ill-posed under many circumstances, with many inverse solutions~\cite{laga2020survey}. In the literature, this problem is referred to as \textit{stereo matching} and has been researched for decades~\cite{lazaros2008review,hamzah2016literature,tippetts2016review,laga2020survey}.

Although stereo vision systems can deliver quality depth maps, 
robust stereo matching searches the left and right images globally for matches, making them computationally expensive~\cite{hamzah2016literature,tippetts2016review}. 
By restricting the range of stereo matching searches, more energy-efficent stereo methods can be developed. Some systems have even been developed which use local matching on FPGAs or ASICs. Despite significant advances~\cite{puglia2017real,lu2021resource}, power/performance trade-offs remain challenging for fully-realized real-time systems.

Recently, deep learning has been explored as an alternative to classical search-based stereo matching methods. Although these methods achieve very promising performance when compared with classical stereo methods, they require at least 100s of MB of RAM and GOPS of compute, ruling them out for low-power edge applications~\cite{laga2020survey}.

\subsection{Depth from Defocus} \label{sec:related-work-dfd}

Depth from Defocus (DfD) is a passive method for generating depth maps from images by analyzing the defocus blurs. This class of method traditionally requires specialized optical modulation in the imaging formation process for engineered defocus, such as phase or amplitude aperture masks~\cite{levin2007image, zhou2011coded, haim2018depth}. Recently, learning-based DfD algorithms have demonstrated the capability to generate high quality depth maps in real time~\cite{haim2018depth, tasneem2022learning, ikoma2021depth}. DfD can be performed using a single camera, an advantage in spatial compactness compared to stereo.  

Recently, depth from differential defocus (DfDD) has emerged as a computationally efficient alternative to traditional DfD methods. Analogous to the computational savings in optic flow from using differential brightness constancy in place of feature tracking, DfDD produces depth maps efficiently by solving closed-form equations on image derivatives caused by differential changes in defocus. DfDD prototypes run at up to 100 FPS, but have still relied on powerful workstations to accomplish this performance~\cite{guo2017focal}. 

Additionally, previous DfDD cameras have handled scene motion in a variety of ways. While early work used motion as its defocus cue~\cite{alexander2016focal,alexander2018focal}, this method proved less stable and more computationally expensive than optically-controlled defocus. Tunable-lens-based DFDD~\cite{guo2017focal,luo2024depth} proved more effective, but due to time-multiplexing the focus changes, are vulnerable to scene motion within paired frames. A metalens-based snapshot camera addresses this issue, but requires custom-fabricated hardware and a narrow bandwidth illumination~\cite{guo2019compact}, limiting its application in practice.


\label{sec:relatedwork}

\section{Methods}
\label{sec:methods}


As detailed in Section~\ref{sec:related-work-dfd}, DfDD is a passive and computationally efficient depth imaging technique, making it attractive for enabling low-power, untethered depth cameras. However, prior DfDD work uses time-division multiplexing to capture differently focused images using the same sensor. This performs poorly under scenes with motion, making it unsuitable for dynamic scenes and cameras that experience ego-motion.

\sysname overcomes this issue by introducing a second image sensor. This allows two differently focused images of the same scene to be captured simultaneously. 
Our contributions include the optomechanical (Section~\ref{subsec:3_optomech}) and mathematical (Section~\ref{subsec:3_1_dfdd}, \ref{subsec:3_2_sensitivity}) developments required to accommodate a second image sensor.

\subsection{Optomechanical Design}
\label{subsec:3_optomech}
As shown in Fig.~\ref{fig:teaser}b, \sysname captures the target scene through a single lens and uses a beamsplitter to guide image copies to two separate sensors, where they are digitized. In order to perform DfDD, the captured images must somehow differ in their focus, which \sysname achieves by 
placing sensors at different optical distances.



Unlike~\cite{guo2019compact}, \sysname's achromatic optomechanics work across the visible spectrum and can be constructed at low-cost with a 3D-printed enclosure and commodity off-the-shelf components, making the creation of \sysname-based systems widely accessible.

\subsection{Depth from Differential Defocus}
\label{subsec:3_1_dfdd}

\begin{figure}
    \centering
    \includegraphics[width=\linewidth]{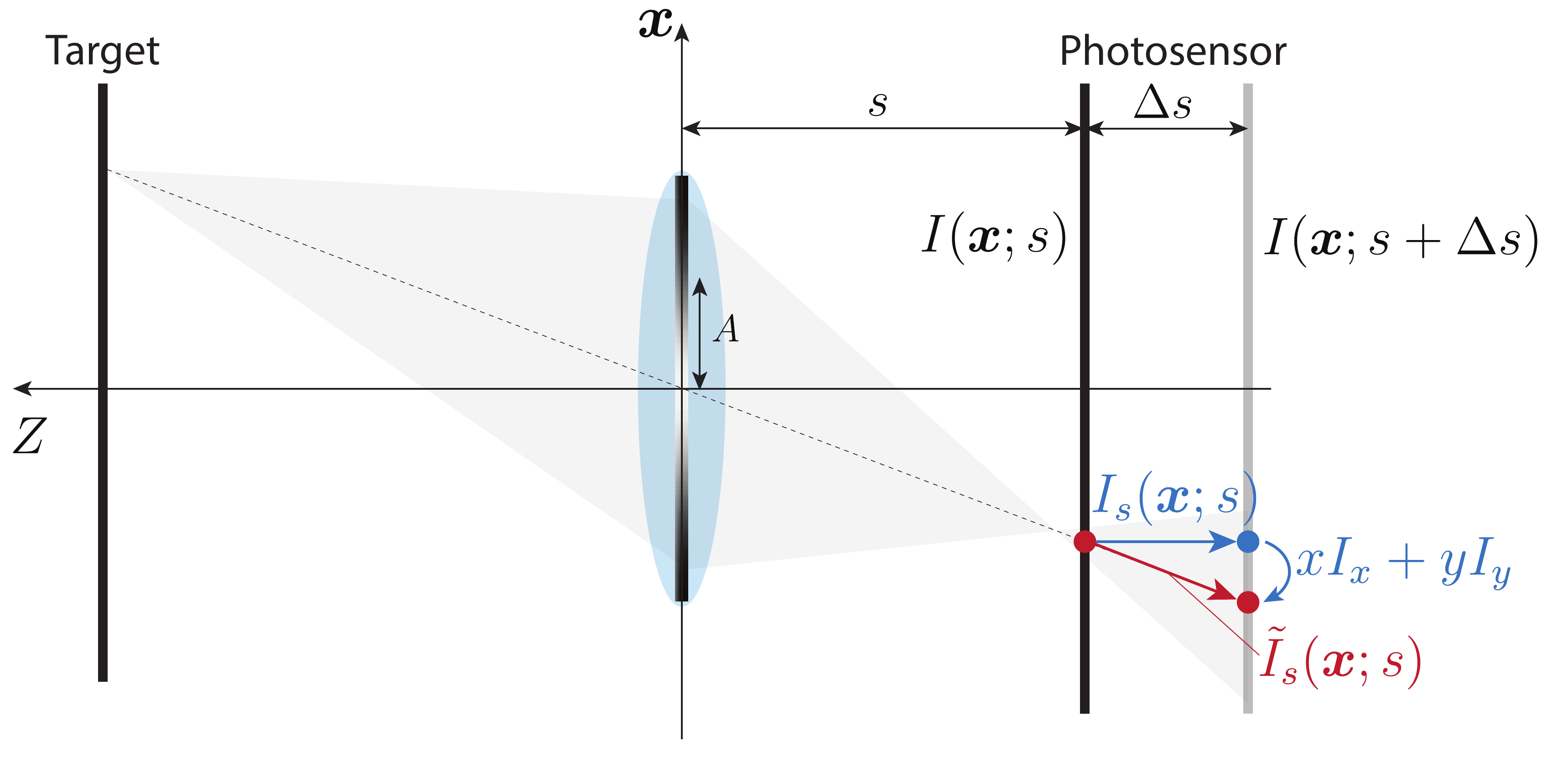}
    \caption{The image formation model. The proposed algorithm calculates the derivative of the aligned images, $\tilde{I}_s$, as a cue for object depth (red arrow). In contrast, previous work~\cite{alexander2019theory} uses two derivatives (blue arrows), $I_s$ and $xI_x+yI_y$, to approximate the same quantity, resulting in higher computation and numerical instability.}
    \label{fig:image-formation}
\end{figure}

As shown in Fig.~\ref{fig:image-formation}, consider a simple scenario in which a front-parallel plane placed at distance $Z$  with texture (spatially-varying brightness) $T$ is imaged through an ideal thin lens with Gaussian blur. Additionally, the distance $s$ between the sensor and the lens, the sensor distance, is allowed to vary, forming an image $I(\boldsymbol{x};s)$. According to the thin-lens model, the image formation process can be described mathematically with a convolution in $\boldsymbol{x}$,
\begin{align}
    I(\boldsymbol{x};s) = k(\boldsymbol{x};s) * P(\boldsymbol{x};s),
    \label{eq:thinlens}
\end{align}
between the all-in-focus pinhole image $P(\boldsymbol{x})$:
\begin{align}
    P(\boldsymbol{x};s) = T\left(-\frac{Z}{s}\boldsymbol{x}\right),
\end{align}
and the Gaussian PSF $k(\boldsymbol{x})$:
\begin{align}
    k(\boldsymbol{x};s) = \frac{1}{\sigma^2} \exp\left(\frac{\lVert \boldsymbol{x}\rVert^2}{2\sigma^2}\right),
\end{align}
where the standard deviation is the defocus level $\sigma$:
\begin{align}
    \sigma = A \left(\frac{1}{Z} - \rho\right)s + A.
    \label{eq:sigma}
\end{align}
The defocus level $\sigma$ is a function of the object distance $Z$, the optical power $\rho$, the lens-to-sensor distance $s$, and the standard deviation of the Gaussian aperture code $A$. 

We note that both the pinhole image $P(\boldsymbol{x};s)$  and the PSF $k(\boldsymbol{x};s)$ depend on the sensor distance $s$. Previous methods account for pinhole magnification change with additional image derivatives (Fig.~\ref{fig:image-formation} blue). We improve robustness and efficiency by correcting the magnification change to isolate the defocus effect. To do this, we register both images to a consensus sensor location $c$. The aligned image $\Tilde{I}$ is generated from a measurement $I(\boldsymbol{x};s)$ by a simple spatial scaling: 

\begin{equation}
    \Tilde{I}(\boldsymbol{x};s) = I\left(\frac{s}{c}\boldsymbol{x}\right).
\end{equation}
The scaled image $\Tilde{I}(\boldsymbol{x};s)$ can be expressed as the 
convolution of the consensus 
pinhole image $P(\boldsymbol{x};c)$, which is independent of $s$, and a scaled PSF $\tilde{k}(\boldsymbol{x};s)$:
\begin{align}
    \tilde{I}(\boldsymbol{x};s) = \tilde{k}(\boldsymbol{x};s) * P(\boldsymbol{x};c).
\end{align}
The scaled PSF $\tilde{k}(\boldsymbol{x};s)$ has the form:
\begin{align}
    \Tilde{k}(\boldsymbol{x};s) &= \left(\frac{s}{c}\right)^2 k\left(\frac{s}{c}\boldsymbol{x}\right),
\end{align}

which has the following mathematical relationship between its derivatives:
\begin{align}
    \tilde{k}_s(\boldsymbol{x};s) = -\frac{c^2 \sigma A}{s^3} \nabla^2 \tilde{k}(\boldsymbol{x};s),
\end{align}
where $\nabla^2$ is the Laplacian in $\boldsymbol{x}$. Because the consensus pinhole image is independent of $s$, the scaled image $\tilde{I}(\boldsymbol{x};s)$ follows a similar relationship:
\begin{equation}
\begin{aligned}
    \Tilde{I}_s(\boldsymbol{x};s) &= \Tilde{k}_s(\boldsymbol{x};s) * P(\boldsymbol{x};c) \\
    &=  -\frac{c^2 \sigma A}{s^3} \nabla^2 \tilde{k}(\boldsymbol{x};s) * P(\boldsymbol{x};c)\\
    &=  -\frac{c^2 \sigma A}{s^3} \nabla^2 \tilde{I}(\boldsymbol{x};s).
\end{aligned}
\end{equation}
Using Eq.~\ref{eq:sigma} for $\sigma$, we obtain the following equation for scene distance $Z$, which applies at every pixel $\boldsymbol{x}$:
\begin{align}
    Z(\boldsymbol{x}) =& \frac{a}{b+\Tilde{I}_s(\boldsymbol{x};s)/\nabla^2 \Tilde{I}(\boldsymbol{x};s)}
    \label{eq:focalsplit},
\end{align}
where $a$ and $b$ are optical constants: 
\begin{align*}
    a =& -A^2, ~ b= -A^2(1/f-1/s).
\end{align*}
For simplicity of notation, we will drop the $(\boldsymbol{x})$ from the equations hereafter whenever the calculation is per-pixel. The full derivation is provided in the supplementary.

Eq.~\ref{eq:focalsplit} can be implemented using the proposed optomechanical setup shown in Fig.~\ref{fig:teaser}c. By capturing a pair of images $I_1$ and $I_2$ at different sensor distances $I_1(\boldsymbol{x}) = I(\boldsymbol{x};s_1)$ and $I_2(\boldsymbol{x}) = I(\boldsymbol{x};s_2)$, we approximate the image derivatives via:
\begin{equation}
\begin{aligned}
    \tilde{I}_{s,\text{approx}} &= I_1(R\boldsymbol{x} + \boldsymbol{t}) - I_2(\boldsymbol{x}), \\
    \nabla^2 \tilde{I}_{\text{approx}} &= \frac{1}{2}\nabla^2 \left(I_1(R\boldsymbol{x} + \boldsymbol{t}) + I_2(\boldsymbol{x})\right),
    \label{eq:approx}
\end{aligned}
\end{equation}
where the matrix $R\in \mathbb{R}^{2\times 2}$ and vector $\boldsymbol{t}\in \mathbb{R}^{2\times 1}$ describe a homography that aligns the images $I_1$ and $I_2$, including the rescaling. 

\subsection{Confidence and sensitivity analysis}
\label{subsec:3_2_sensitivity}



\noindent\textbf{Confidence.} Like other DfDD algorithms, our new depth equation (Eq.~\ref{eq:focalsplit}) generates a solution even in textureless or blurred-out regions, but the lack of image contrast will cause numerical instability in the ratio of small derivatives $\tilde{I}_s$ and $\nabla^2\tilde{I}$. 
Fortunately, this degeneracy can be predicted with confidence from the measured values of $\tilde{I}_s$. As shown from the simulation result in Fig.~\ref{fig:sensitivity}, the overall depth estimation error is approximately inversely proportional to the magnitude of the image derivative $|\tilde{I}_s|$. Thus, we can define a simple confidence metric at each pixel:
\begin{align}
    C = \tilde{I}_s^2,
    \label{eq:conf}
\end{align}
and use the confidence value $C$ to filter out depth predictions according to a preset threshold $C_{\text{thre}}$. Sec.~\ref{sec:results} shows the effectiveness of the confidence metric in the real data. 


\noindent\textbf{Working range.} As objects become farther away from the plane of focus, the defocus blur gradually attenuates the intensity variations in the images. Meanwhile, as the image noise stays constant, the signal-to-noise ratio (SNR) of both image derivatives $\tilde{I}_s$ and $\nabla^2\tilde{I}$ gradually reduces. Here, we mathematically define the SNR of $\tilde{I}_s$ and $\nabla^2\tilde{I}$ as:
\begin{equation}
    \begin{aligned}
        \text{SNR}(\tilde{I}_s) &= \frac{\tilde{I}_s}{|\tilde{I}_s - \tilde{I}_{s,\text{approx}}|}, \\
        \text{SNR}(\nabla^2 \tilde{I}) &= \frac{\tilde{I}_s}{|\nabla^2 \tilde{I} - \nabla^2 \tilde{I}_{\text{approx}}|},
    \end{aligned}
\end{equation}
where the subscript $\text{approx}$ indicate the approximated derivatives using Eq.~\ref{eq:approx}. Fig.~\ref{fig:sensitivity}b validates the significant decrease in SNR of $\tilde{I}_s$ and $\nabla^2\tilde{I}$ as the object departs from the plane of focus (dashed line.) 

This evidence suggests that the proposed method has a natural \textit{working range}, i.e., a region around the plane of focus where the algorithm's prediction is accurate. Empirically, we determine the working range as the depth region with a mean depth prediction error smaller than $5\%$ of the true depth. 

\noindent\textbf{Numerical accuracy.} The new depth equation (Eq.~\ref{eq:focalsplit}) outperforms the previously suggested equation for a layered-retina system~\cite{alexander2019theory}, where $Z$ was directly calculated from derivatives of the un-aligned images $I(\boldsymbol{x};s)$ via:
\begin{align}
    Z =& \frac{a }{b + \left(d (x I_x + y I_y) + I_s\right)/ \nabla^2 I}, \label{eq:old}
\end{align}
with the additional constant $d = 1/s^2$ on a magnification term $xI_x + yI_y$ that our method does not need to compute. As illustrated in Fig.~\ref{fig:image-formation}, the two methods, Eq.~\ref{eq:focalsplit} and Eq.~\ref{eq:old}, differ in 
how they account for brightness changes across sensor distances. By rescaling the images, the proposed method removes the effect of the magnification term $xI_x + yI_y$ and loosens the requirement that the sensor distance change is differential in scale. The proposed method (Eq.~\ref{eq:focalsplit}) is more robust to larger variations in magnification than Eq.~\ref{eq:old}. This can be shown by the simulation analysis in Fig.~\ref{fig:sensitivity}c, where we study the depth accuracy as a function of sensor distance variation, $\Delta s = |s_1 - s_2|$. In the absence of noise, both methods show a rise in depth prediction error when $\Delta s$ increase, but the proposed algorithm (Eq.~\ref{eq:focalsplit}) consistently achieves a lower error than the prior one (Eq.~\ref{eq:old}).


\begin{figure}
    \centering
    \includegraphics[width=\linewidth]{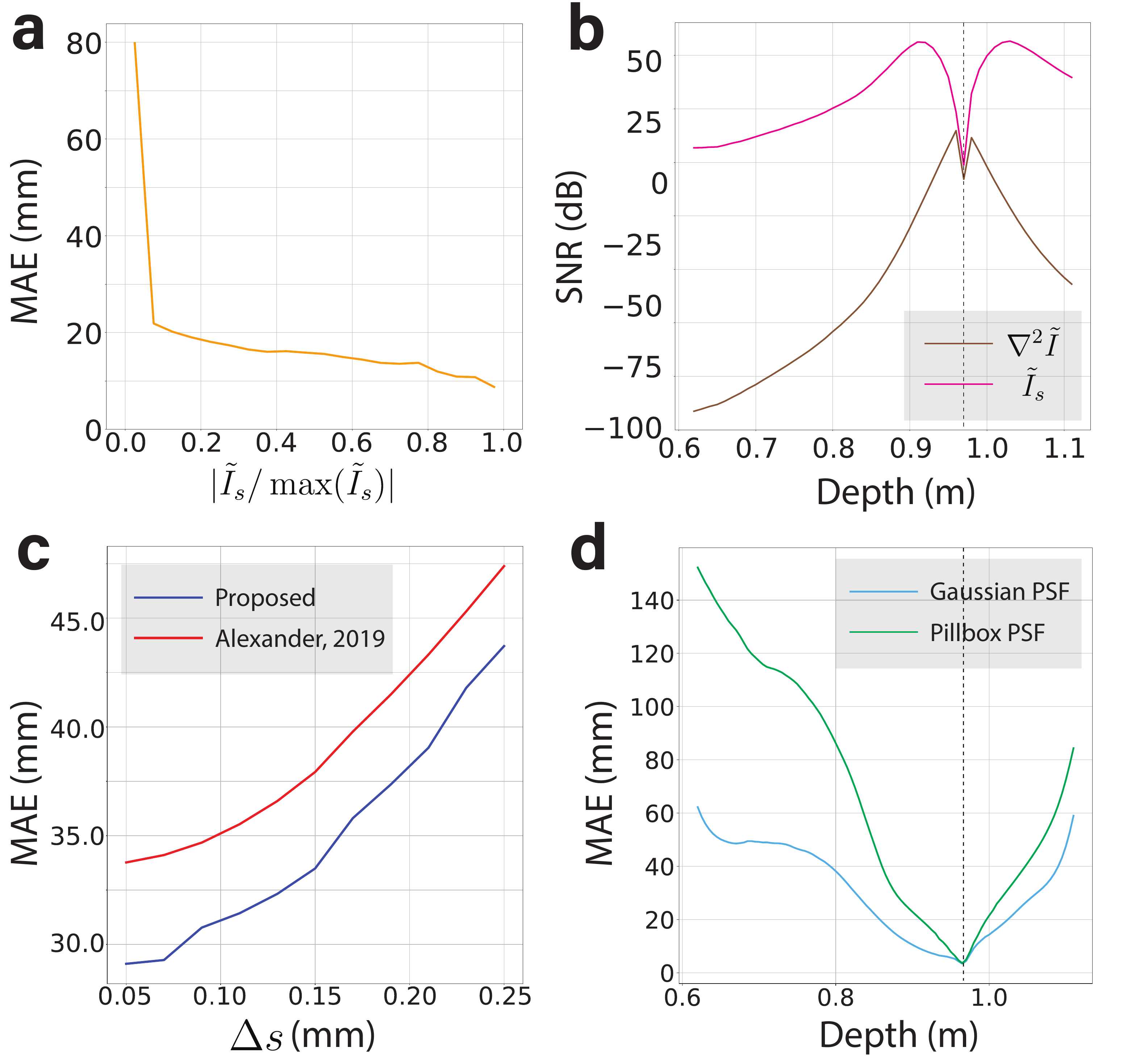}
    \caption{Sensitivity analysis using synthetic data. We simulate the image pair, $I_1$ and $I_2$, of front parallel textured planes placed at different depths $Z$ and use the data to analyze the sensitivity of the algorithm. (a) Validation of the confidence metric. The overall depth estimation error, quantified by the mean absolute error (MAE), monotonically decreases as the normalized image derivative $|\tilde{I}_s/\max(\tilde{I}_s)|$ increases, suggesting the latter to be an effective confidence metric of the depth prediction. (b) Signal-to-noise ratio (SNR) of the estimated image derivatives $\tilde{I}_s$ and $\nabla^2\tilde{I}$ from finite difference. The vertical dashed line indicates the depth of the focal plane. The depth estimation becomes noisy when the SNR of both derivatives is too low. (c) Overall depth estimation error of using the proposed depth equation (Eq.~\ref{eq:focalsplit}) vs. the previously suggested equation (Eq.~\ref{eq:old}). The proposed one universally achieves higher accuracy for all sensor distance variation $\Delta s$. (d) Depth estimation error for Gaussian and Pillbox-shaped PSFs. }
    \label{fig:sensitivity}
    \vspace{-0.1in}
\end{figure}

\noindent\textbf{Aperture code.} The derivation of Eq.~\ref{eq:focalsplit} requires Gaussian PSFs. However, in practice, most cameras have a disk aperture, which leads to pillbox-shaped PSFs. Fig.~\ref{fig:sensitivity}d shows an increase in depth prediction error if directly using images captured with pillbox PSFs in simulation. However, it indicates the proposed algorithm can be directly applied to pillbox PSFs if the error can be tolerated.

\section{Prototype System}
\label{sec:system}

Utilizing the proposed depth sensing algorithm (Eq.~\ref{eq:focalsplit}), we design and build a depth camera with fully onboard power and compute using off-the-shelf optics, image sensors, and a single-board computer housed in a custom 3D-printed enclosure. The system can be powered by any 5VDC power bank, making it untethered and handheld. We include a complete DIY guide in the supplementary with a list of parts, the 3D-printing model, and assembly and calibration instructions to rebuild the prototype with $\$500$ budget. As a high-level summary, the system (Fig.~\ref{fig:teaser}c) consists of a 30 mm lens, a non-polarizing cube beamsplitter, and two OV5647 RGB image sensors connected to a Raspberry Pi 5 to perform image capture and data processing. The two photosensors are designed to have a 0.4 mm difference in sensor distances.

\begin{figure}
    \centering
    \includegraphics[width=\linewidth]{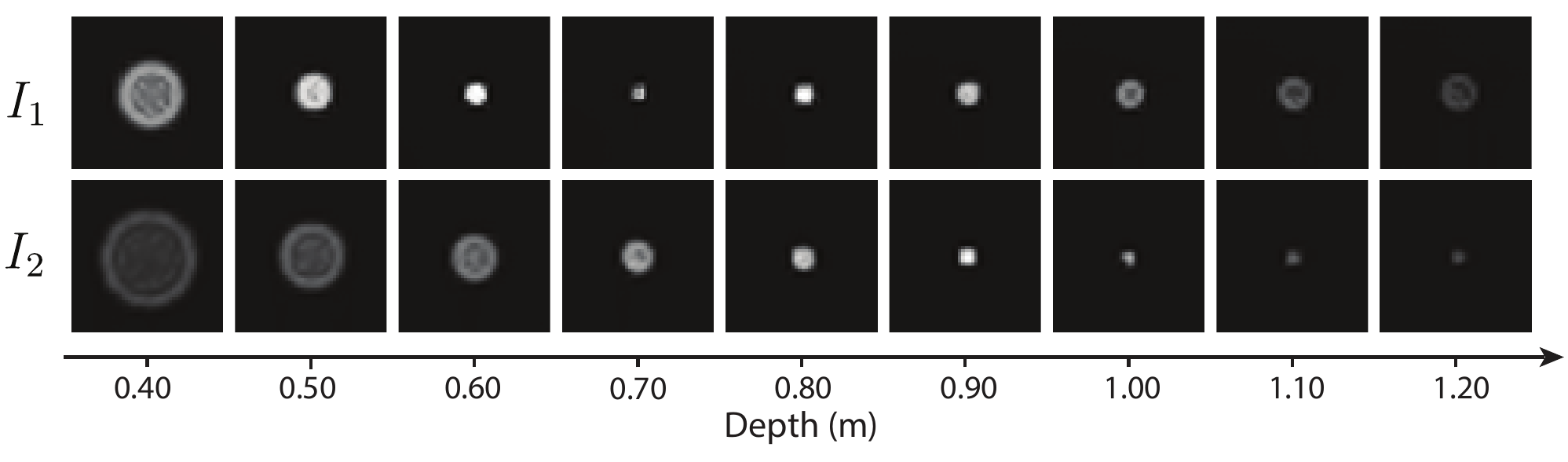}
    \caption{PSFs of the image pair, $I_1$ and $I_2$, at different depths using the assembled Focal Split prototype. The PSFs are measured by taking pictures of a white LED point source. The focal planes of $I_1$ and $I_2$ are approximately at 0.7 m and 1.2 m, respectively.}
    \label{fig:psfs}
\end{figure}

\subsection{Implementation}

After simultaneously capturing the two images, $I_1^{\text{RGB}}$ and $I_2^{\text{RGB}}$, from the two photosensors. We convert them to grayscale images, $I_1$ and $I_2$, for the depth estimation. 

\noindent\textbf{Aberration correction.} First, we reduce the non-uniform background lighting in $I_1$ and $I_2$, as it severely contaminates the approximation of the image derivatives $\tilde{I}_s$ using Eq.~\ref{eq:approx}. We adopt the same practice as prior works~\cite{guo2019compact,luo2024depth} to filter out the background lighting after measuring the images: 
\begin{align}
    I_{i}^{\text{bck}} = I_i - \frac{1}{K^2} B * I_i, \qquad i = 1,2,
    \label{eq:highpass}
\end{align}
where $B$ is a K$\times$K box filter. We set $K=21$ in this work based on our experience.

\noindent\textbf{Noise attenuation.} The photosensor we use has a significant camera noise. Thus, we apply a Gaussian filter to suppress the noise in the measurements after the aberration correction:
\begin{align}
    I_{i}^{\text{clean}} = G * I_i^{\text{bck}}, \qquad i = 1,2.
    \label{eq:gaussian}
\end{align}
In our experiment, we set the standard deviation of the Gaussian filter $G$ to be 11 pixels.


\noindent\textbf{Alignment.} We determine the homography between the two photosensors from the corresponding SIFT key points from the two images. Then, we compute the approximated image derivatives, $\nabla^2\Tilde{I}$ and $\Tilde{I}_s$, via Eq.~\ref{eq:approx}.

\noindent\textbf{Depth estimation.} Although our depth estimation algorithm (Eq.~\ref{eq:focalsplit}) can be performed per pixel, we aggregate the image derivatives $\tilde{I}_s$ and $\nabla^2 \tilde{I}$ within an image patch when computing the depth value to attenuate the noise:
\begin{align}
    Z(\boldsymbol{x}) =& \frac{a \nabla^2 \Tilde{I}(\boldsymbol{x};s) \left(b \nabla^2 \Tilde{I}(\boldsymbol{x};s) +\Tilde{I}_s(\boldsymbol{x};s)\right) * W }{\left(b \nabla^2 \Tilde{I}(\boldsymbol{x};s) +\Tilde{I}_s(\boldsymbol{x};s)\right)^2 * W},
    \label{eq:window}
\end{align}
where the L$\times$L box filter $W$. We set $L=21$ in our implementation.

\noindent\textbf{Parameter calibration.} The only parameters that need calibration in the entire implementation are the $a$ and $b$ in Eq.~\ref{eq:window}. We utilize a data-driven approach to determining their values. By moving a front-parallel texture to a series of distances $\{Z^*_j, j=1,\cdots,J\}$ and capturing an image pair for each distance $\{I_{i,j}, i=1,2, j=1,\cdots,J\}$, and optimizing the following objective function, we obtain the calibrated parameters $a$ and $b$:
\begin{align}
    \arg\min_{a,b} \sum_{j=1}^J\lVert Z^*_j - Z(\boldsymbol{x};I_{1,j},I_{2,j}, a, b) \rVert^2.
\end{align}







\section{Real-World Results}
\label{sec:results}

\begin{figure}
    \centering
    \includegraphics[width=\linewidth]{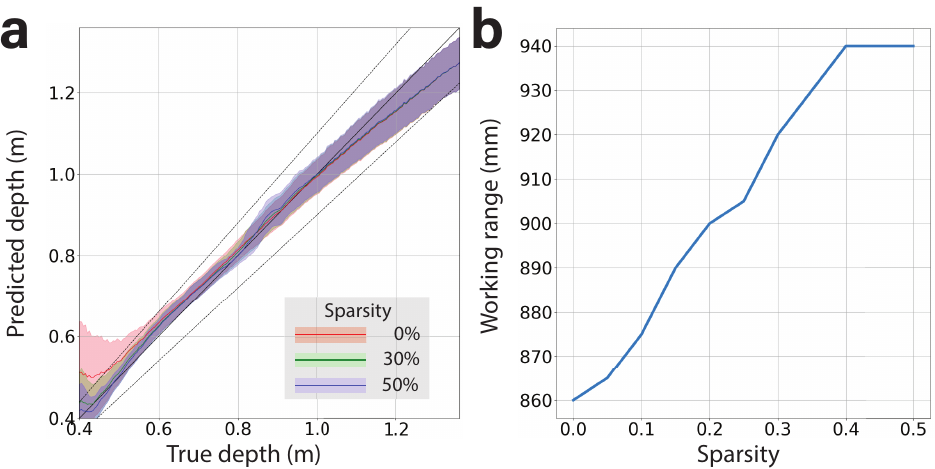}
    \caption{Quantitative analysis of the Focal Split prototype using real captured data. (a) Depth estimation accuracy at different confidence levels. The sparsity indicates the percentage of discarded, least-confident pixels. (b) Working range, defined as the depths where the MAE is smaller than $5\%$ of the true depth, as a function of confidence levels. 
    }
    \label{fig:quant}
\end{figure}

\begin{table*}[h!]
\vspace{-0.3in}
\caption{System level comparison of monocular passive depth imaging techniques. Only ours achieved untethered depth estimation.}
\vspace{-0.1in}
\label{tab:system-compare}
\centering
\footnotesize
\begin{tabular}{lccccccc}
\hline
\textbf{Name}                                     & \textbf{Technique}   & \textbf{\begin{tabular}[c]{@{}c@{}}\# Sequential \\ Capture\end{tabular}} & \textbf{\begin{tabular}[c]{@{}c@{}}Depth Map \\ Resolution\end{tabular}}                         & \textbf{\begin{tabular}[c]{@{}c@{}}Dense \\ Depth Map?\end{tabular}} & \textbf{Untethered?} & \textbf{Real Time?} & \textbf{Platform / Power}                                                                                                \\ \hline
Newcombe~\cite{newcombe2011dtam}                  & SfM                  & $>$10                                                                     & 640$\times$480                                                                                   & \cmark                                                               & \xmark               & \cmark              & \begin{tabular}[c]{@{}c@{}}NVIDIA GTX 480 GPU\\ i7 quad-core CPU\end{tabular}                                            \\ \hline
Schonberger~\cite{schonberger2016structure}       & SfM                  & $\sim$10,000                                                              & Variable                                                                                         & \xmark                                                               & \xmark               & \xmark              & \begin{tabular}[c]{@{}c@{}}2.7 GHz Processor\\ 256GB RAM\end{tabular}                                                    \\ \hline
\multirow{2}{*}{Tang et al.~\cite{tang2017depth}} & \multirow{2}{*}{DfD} & \multirow{2}{*}{2}                                                        & \multirow{2}{*}{\begin{tabular}[c]{@{}c@{}}5184$\times$3456 or \\ 2464$\times$3280\end{tabular}} & Initial: \xmark                                                      & \xmark               & \cmark              & \multirow{2}{*}{\begin{tabular}[c]{@{}c@{}}Two 8-Core 2.6 GHz \\ Xeon CPU 128 GB RAM\end{tabular}}                       \\ \cline{5-7}
                                                  &                      &                                                                           &                                                                                                  & Refine: \cmark                                                       & \xmark               & \xmark              &                                                                                                                          \\ \hline
Focal Flow~\cite{alexander2018focal}              & DfDD                 & 3                                                                         & 960$\times$600                                                                                   & \xmark                                                               & \xmark               & \cmark              & \begin{tabular}[c]{@{}c@{}}2.93 GHz Xeon \\ X5570 CPU\end{tabular}                                                       \\ \hline
Haim et al.~\cite{haim2018depth}                  & DfD                  & 1                                                                         & 1920$\times$1080                                                                                 & \cmark                                                               & \xmark               & \xmark              & \begin{tabular}[c]{@{}c@{}}Nvidia Titan X \\ Pascal GPU\end{tabular}                                                     \\ \hline
Ikoma et al.~\cite{ikoma2021depth}                & DfD                  & 1                                                                         & 384$\times$384                                                                                   & \cmark                                                               & \xmark               & \cmark              & \begin{tabular}[c]{@{}c@{}}Unspecified Platform\\ 124 kFLOPs/pixel\end{tabular}                                          \\ \hline
Focal Track~\cite{guo2017focal}                   & DfDD                 & 2                                                                         & 480$\times$300                                                                                   & \xmark                                                               & \xmark               & \cmark              & \begin{tabular}[c]{@{}c@{}}NVIDIA GeForce GTX 1080 \\ Notebook Graphics Card \\ Intel Core i7 6820HK CPU\end{tabular}    \\ \hline
COD~\cite{luo2024depth}                           & DfDD                 & 4                                                                         & 480$\times$300                                                                                   & \xmark                                                               & \xmark               & \xmark              & \begin{tabular}[c]{@{}c@{}}Intel Core i9-11900K \\ Processor\end{tabular}                                                \\ \hline
Ours                                              & DfDD                 & 1                                                                         & 480$\times$360                                                                                   & \xmark                                                               & \cmark               & \cmark              & \begin{tabular}[c]{@{}c@{}}Raspberry Pi 5 with 2.4 GHz \\ ARM Cortex-A76 Processor\\ 500 FLOPs/pixel, 4.9 W\end{tabular} \\ \hline
\end{tabular}
\vspace{-0.1in}
\end{table*}

\subsection{Quantitative Analysis}

\noindent\textbf{Working range and depth accuracy.} We analyze the working range and depth estimation accuracy using front-parallel textured planes placed at a series of known distances. Fig.~\ref{fig:quant}a-b visualizes the depth estimation accuracy and working range of the \sysname prototype under different confidence thresholds. The confidence thresholds are determined by the overall sparsity of the remaining pixels. By setting the sparsity to $40\%$, i.e., discarding the $40\%$ least confident pixels, the working range increases from 860 mm to 940 mm, demonstrating the effectiveness of the simple confidence metric.

\noindent\textbf{System characteristics.} Table~\ref{tab:system-compare} analyzes the specifications of different passive monocular depth estimation systems. Due to the diverse camera setups and computing platforms of these methods, it is challenging to directly and fairly compare these methods in terms of power consumption. Instead, we list the number of sequential captures, depth map resolutions, frame rates, and computing platforms of each method. Compared to methods that generate dense depth maps on more power-hungry platforms, Focal Split provides a complementary option to produce sparse depth maps with low power consumption.

Our Focal Split prototypes generate a 480$\times$360 depth map every 0.47 sec at a power consumption of 4.9 W by running an un-optimized Python script, within which 0.10 sec were used for image measurement and the remaining 0.37 sec were for data processing. Considering the low FLOPs per pixel, the frame rate can be significantly improved if the script is pre-compiled or multi-threading is used. Furthermore, the power consumption of an idling Raspberry Pi is 2.5 W. 



\noindent\textbf{Comparison with other DfD algorithms.} Focal Split's critical advantage compared to previous computationally efficient DfD algorithms is its snapshot capability. We quantitatively analyze this advantage in Tab.~\ref{tab:mae}. We adopt the front parallel textured planes as the scene and use the Focal Split prototype to resemble several other DfD systems, i.e., Focal Flow~\cite{alexander2018focal}, Tang et al.~\cite{tang2017depth}, and Focal Track~\cite{guo2017focal}. Focal Flow requires three input images with relative motion in between. Thus, we use our prototype to capture three $I_1$ consecutively while the target is translated axially by 5 mm between the adjacent measurements. Focal Track uses two sequentially measured images of a static scene with different focal planes. Thus, we measure $I_1$ and $I_2$ at different time stamps and align them using precalibrated homography as the input data. We deliberately move the target when capturing $I_1$ and $I_2$ to resemble a dynamic scene. We use the same input data as Focal Track for Tang et al. Table~\ref{tab:mae} clearly shows the degradation of depth estimation accuracy of these methods when the scene is dynamic, while Focal Split's accuracy remains invariant under noise. 



\begin{table}[h]
\caption{Quantitative comparison between the proposed method and other computationally efficient DfD algorithms using real data. We use the Focal Split prototype to capture the required data for each method. When the scene is static, Focal Track's algorithm is effectively equivalent to ours. Thus, both methods have the same accuracy and working range. However, when the target is dynamic, Focal Track and Tang et al. significantly degrade, while ours remains constant thanks to its snapshot functionality. Focal Flow is designed for dynamic scenes, but its accuracy and working range are both worse than ours. See details about the data collection in Sec.~\ref{secsec:depthresults}.}
\centering
\label{tab:mae}
\resizebox{\linewidth}{!}{
\begin{tabular}{l|cc|cccc}
\hline
\multirow{3}{*}{\textbf{Method}} 
& \multicolumn{2}{c|}{\textbf{Static Scenes}} 
& \multicolumn{2}{c}{\textbf{Dynamic Scenes}} \\
\cline{2-5}
& \textbf{MAE} & \textbf{Working Range} 
& \textbf{MAE} & \textbf{Working Range} \\ 
& \textbf{(mm)} & \textbf{(m)}
& \textbf{(mm)} & \textbf{(m)} \\
\hline
Focal Flow~\cite{alexander2018focal}      & - & -   & 179.25   & 0.600 \\
Tang et al.~\cite{tang2017depth}             & 109.97          & 0.355  & 316.78 & 0.145\\
Focal Track~\cite{guo2017focal}              & 41.82  & 0.860  &  107.69 & 0.295 \\
\textbf{Ours}                                 & \textbf{41.82}  & \textbf{0.860} & \textbf{41.82} & \textbf{0.860} \\
\hline
\end{tabular}
}
\end{table}











\begin{figure*}[h!]
\vspace{-0.2in}
    \centering
    \includegraphics[width=1.0\linewidth]{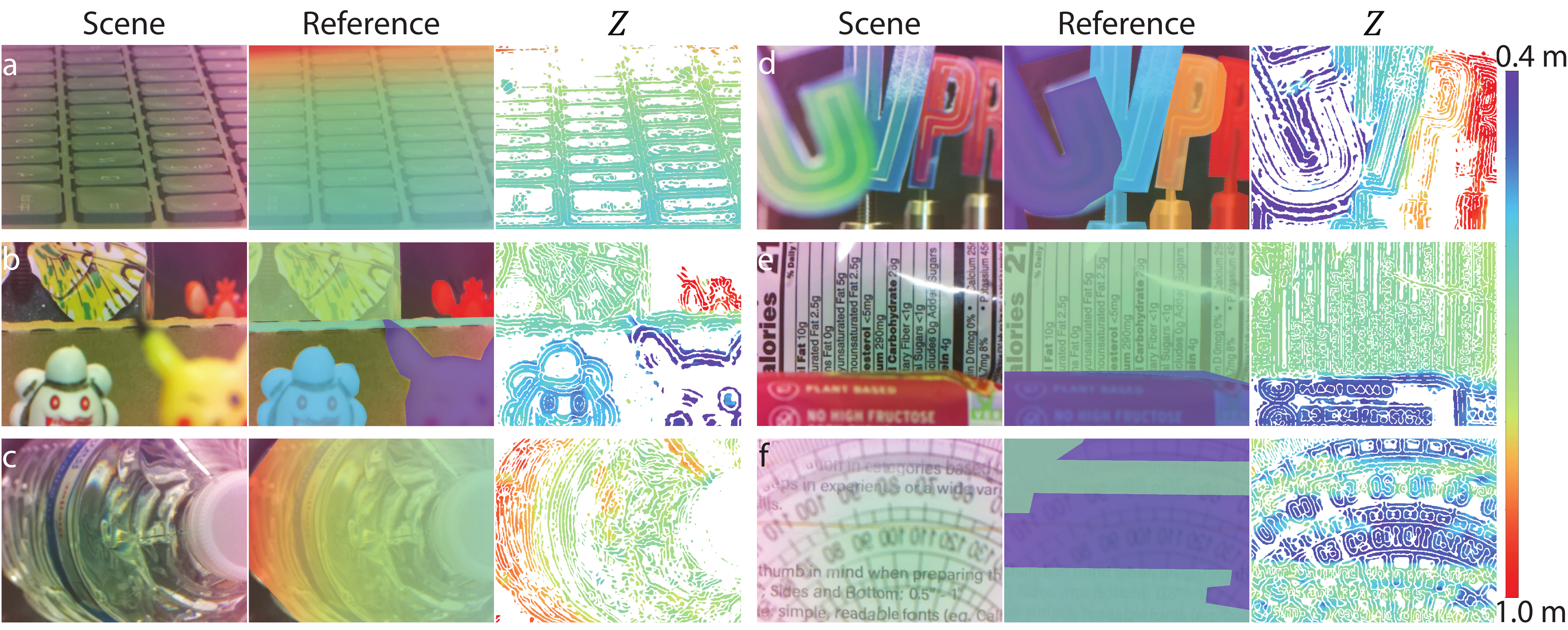}
    \caption{Sample depth maps captured by \sysname. The depth maps are filtered by the confidence metric with a constant confidence threshold $C_{\text{thre}}$. The reference depth maps are manually measured to provide a qualitative evaluation. More results are in the supplementary.
    }
    \label{fig:depthmap}
\end{figure*}

\begin{figure*}[h!]
    \centering
    \includegraphics[width=0.9\linewidth]{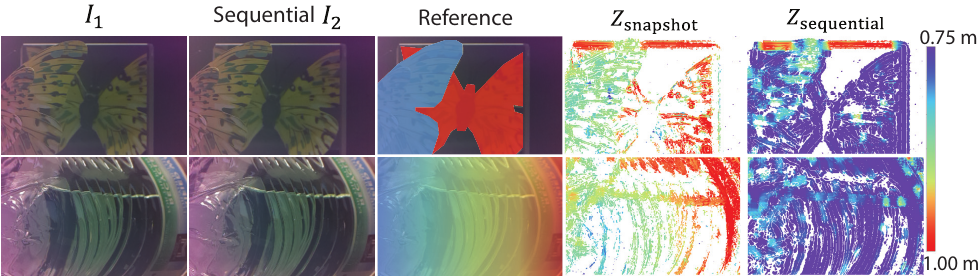}
    \caption{Depth estimation of dynamic scenes using snapshot vs. sequentially measured $I_1$ and $I_2$. The symbols $Z_{\text{snapshot}}$ and $Z_{\text{sequential}}$ represent the depth maps generated from each capture method, respectively. The depth maps are visualized with the same confidence threshold. The depth map is completely contaminated in $Z_{\text{sequential}}$, caused by the displacement of objects in the two frames due to sequential measurement. Furthermore, these artifacts cannot be identified and removed by the confidence metrics. This experiment demonstrates the critical advantage of snapshot measurement for depth from differential defocus. }
    \label{fig:motion}
\end{figure*}

\subsection{Depth Maps}
\label{secsec:depthresults}

Fig.~\ref{fig:depthmap} shows sample depth maps generated by the \sysname prototype. The confidence effectively filters out unreliable depth predictions. Besides traditional textured objects, our method can utilize any intensity changes in the images, not necessarily textures, as cues to measure depth. Fig.~\ref{fig:depthmap}c demonstrates using caustics in the water bottle to estimate the depth values, which could be challenging for stereo-based solutions as the caustics are view-dependent. Fig.~\ref{fig:depthmap}f demonstrates an interesting scenario where the foreground, a protractor, is semi-transparent. In this scenario, traditional dense depth maps are insufficient to represent the scene structure. At each pixel, Focal Split outputs the depth values of the surface with stronger textures in the area centered around it, allowing simultaneous depth estimation for both foreground and background. Meanwhile, a current limitation of Focal Split is that depth estimation becomes inaccurate when both foreground and background have strong textures. Future work includes the prediction of such regions and independently predicting the foreground and background depth values. More depth maps can be found on the project page listed in the abstract.

Besides, we also compare the depth map quality when the input images, $I_1$ and $I_2$, are captured in a snapshot vs. sequentially when the target is displaced. As shown in Fig.~\ref{fig:motion}, the sequential measurement clearly results in contaminated depth map measurements compared to snapshot measurements.

\vspace{-0.1in}
\paragraph{Acknowledgement.} This research was partially supported by the National Science Foundation under awards numbers CNS-2145584, CNS-2400463, and CIF-2431505. We would also like to acknowledge support by the Alfred P. Sloan Foundation,  VMWare, and Catherine M. and James E. Allchin. Any opinions, findings, conclusions, or recommendations expressed in this material are those of the authors and do not necessarily reflect the views of the National Science Foundation or other supporters.

\newpage 

\label{sec:discussion}


{
    \small
    \bibliographystyle{ieeenat_fullname}
    \bibliography{main}
}

 \clearpage
\setcounter{page}{1}
\maketitlesupplementary

\section{Depth from Differential Defocus}

This section provides additional clarifications that complement the derivation of the proposed depth estimation algorithm (Eq. 11 in the main paper). For the convenience of reading, we list all key quantities and symbols already introduced in the main paper in Table~\ref{tab:equation_review}.

Consider the image captured with sensor distance $s$ of a target with object distance $Z$, $I(\boldsymbol{x};s)$. As discussed in the main paper, the variation of the sensor distance $s$ will change both the image defocus and the magnification. Thus, we rescale the image $I(\boldsymbol{x};s)$ via:
\begin{equation}
    \Tilde{I}(\boldsymbol{x};s) = I\left(\frac{s}{c}\boldsymbol{x}\right),
    \label{eq:mag_model}
\end{equation}
and the aligned image $\Tilde{I}(\boldsymbol{x};s)$ has a constant magnification independent of the sensor distance $s$. 

As the unaligned images $I(\boldsymbol{x};s)$ is the convolution of the point spread function (PSF) $k(\boldsymbol{x};s)$ and the pinhole image $P(\boldsymbol{x};s)$, as shown in Table~\ref{tab:equation_review}, the aligned images $\Tilde{I}(\boldsymbol{x};s)$ satisfies:
\begin{equation}
    \begin{split}
        &\Tilde{I}(\boldsymbol{x};s) = \iint k\left( \frac{s}{c}(\boldsymbol{x} - \boldsymbol{u});s \right)  P\left( \frac{s}{c} \boldsymbol{u};s \right) \frac{s^{2}}{c^2} |d\boldsymbol{u}| \\
        &= \iint \frac{s^{2}}{c^2} k\left( \frac{s}{c}(\boldsymbol{x} - \boldsymbol{u});s \right) T\left(-\frac{Z}{c}\boldsymbol{u}\right) |d\boldsymbol{u}|.
    \end{split}
    \label{eq:expanded_mag_model}
\end{equation}
Thus, if we define the scaled PSF to be:
\begin{align}
        \Tilde{k}(\boldsymbol{x};s)=\frac{s^{2}}{c^2}k(\frac{s}{c}\boldsymbol{x};s),
    \label{eq:k_star}
\end{align}
and the consensus pinhole image as:
\begin{equation}
    \Tilde{P}(\boldsymbol{x}; c) = P\left( \frac{s}{c} \boldsymbol{x};s \right),
    \label{eq:P_star}
\end{equation}
the aligned image $\Tilde{I}(\boldsymbol{x};s)$ is the convolution of the two:
\begin{equation}
    \Tilde{I}(\boldsymbol{x}) = \Tilde{k}(\boldsymbol{x};s) * \Tilde{P}(\boldsymbol{x};c).
    \label{eq:mag_model_final}
\end{equation}

\begin{table}
    \centering
    \renewcommand{\arraystretch}{1.2}
    \begin{tabular}{|>{\centering\arraybackslash}S{m{1cm}}|>{\centering\arraybackslash}S{m{3cm}}|m{3.2cm}|}
        \hline
        \textbf{Symbol} & \textbf{Expression} & \textbf{Description} \\
        \hline
        $T\left(\boldsymbol{x}\right)$ & - & Object texture \\
        \hline
        $P(\boldsymbol{x};s)$ & $T\left(-\frac{Z}{s}x,-\frac{Z}{s}y\right)$ & Pinhole image \\
        \hline
        $k(\boldsymbol{x};s)$ & $\frac{1}{\sigma^2(s)} ~ \exp\left(-\frac{x^2+y^2}{2\sigma(s)^2}\right)$ & Point spread function (PSF) \\
        \hline
        $I(\boldsymbol{x};s)$ & $k(\boldsymbol{x};s) * P(\boldsymbol{x};s)$ & Defocused image \\
        \hline
        $\sigma(s)$ & $A(\frac{1}{Z} - \rho)s + A$ & Defocus level \\
        \hline
        $Z$ & - & Object depth \\
        \hline
        $s$ & - & Sensor distance\\
        \hline
        $A$ & - & Standard deviation of the Gaussian aperture code \\
        \hline
        $\rho$ & - & Optical power of the lens \\
        \hline
    \end{tabular}
    \caption{List of key symbols and quantities for deriving the proposed depth estimation algorithm.}
    \label{tab:equation_review}
\end{table}

The relationship between the partial derivatives of the scaled PSFs $\Tilde{k}(\boldsymbol{x};s)$, shown in Eq.~9 in the main paper, can be derived as follows. The partial derivative of $\Tilde{k}(\boldsymbol{x};s)$ with respect to the sensor distance $s$ is:
\begin{equation}
    \begin{split}
        \Tilde{k}_s(\boldsymbol{x};s) = (2+2\psi) \frac{sA}{c^2\sigma^3} e^{\psi},
    \end{split}
    \label{eq:I_s}
\end{equation}
where 
\begin{equation}
    \psi = -\frac{s^2\lVert \boldsymbol{x} \rVert^2}{2c^2 \sigma^2}.
    \label{eq:I_s}
\end{equation}
And the spatial Laplacian of $\Tilde{k}(\boldsymbol{x};s)$ is:
\begin{equation}
    \nabla^2 \Tilde{k}(\boldsymbol{x};s) =  -(2+2\psi) \frac{s^4}{c^4\sigma^4} e^{\psi}.
    \label{eq:lapI}
\end{equation}
By combining Eq.~\ref{eq:I_s}~and~\ref{eq:lapI}, we obtain the relationship:
\begin{align}
    \tilde{k}_s(\boldsymbol{x};s) = -\frac{c^2 \sigma A}{s^3} \nabla^2 \tilde{k}(\boldsymbol{x};s),
\end{align}

\section{\textit{Do It Yourself} Guide}

\subsection{Components}

The~\sysname prototype was assembled mainly using off-the-shelf optical and mechanical components. We provide the complete list of parts in Table~\ref{tab:component}. We built the custom housing via 3D-printing and the CAD model is included in the project page listed in the abstract.

\begin{figure}[h!]
    \centering
    \includegraphics[width=\linewidth]{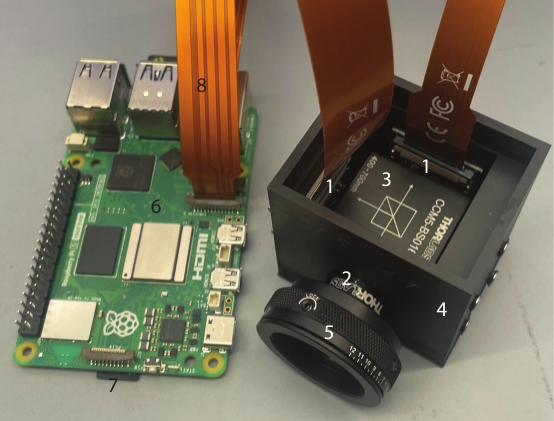}
    \caption{\sysname prototype. }
    \label{fig:box}
\end{figure}

\begin{table*}[h!]
    \centering
    \renewcommand{\arraystretch}{1.2}
    \begin{tabular}{|c|p{2cm}|c|c|c|c|p{5cm}|}
        \hline
        \textbf{No.} & \textbf{Component} & \textbf{Vendor} & \textbf{Part Number} & \textbf{Quantity} & \textbf{Price} & \textbf{Description} \\
        \hline
        1 & Photosensor & PiShop & 799632837664 & 2 & \$31 & Low-power RGB sensor, with the lens removed \\
        \hline
        2 & Lens & Thorlabs & AC127-030-A-ML & 1 & \$93 & Ø1/2" Mounted Achromatic Doublets lens with AR coated \\
        \hline
        3 & Beamsplitter & Thorlabs & CCM5-BS016/M & 1 & \$210 & Cube-Mounted, Non-Polarizing, 50:50 Beamsplitter Cube \\
        \hline
        4 & Housing &  &  & 1 & - & Housing to hold sensors, lens, and beamsplitter, produced by 3D printing \\
        \hline
        5 & Iris & Thorlabs & SM1D12C & 1 & \$120 & (Optional) Manual iris to control the brightness of captured images \\
        \hline
        6 & Raspberry Pi & PiShop & 4GB-9024 & 1 & \$60 & Computing unit \\
        \hline
        7 & MicroSD Card & PiShop & 1337-1 & 1 & \$13 & SD card for the computer \\
        \hline
        8 & Camera Cable & PiShop & 1510-1 & 2 & \$2 & Cable for sensors \\
        \hline
        9 & Power Supply & PiShop & 1795-1 & 1 & \$14 & Power Supply for the computer \\
        \hline
        10 & LCD & PiShop & 1824 & 1 & \$23 & (Optional) Display used for showing depth maps and photos \\\hline
        11 & 5 VDC Battery Pack & Thorlabs & CPS1 & 1 & \$41 & Battery \\
        \hline
    \end{tabular}
    \caption{List of parts.}
    \label{tab:component}
\end{table*}


\subsection{Assembling Instruction}

\subsubsection{Opto-mechanical System}


First, disassemble the mounted lenses from the photosensors (1), and connect the camera cable (8) to each photosensor. Use M2 screws and washers to mount the two photosensor circuits onto the housing (4). Make sure to add an extra washer when mounting one of the sensors so that its sensor distance is different from the other one. Then, mount the beamsplitter (3) onto the housing using M4 screws and washers. Be mindful that its orientation is the same as shown in Fig.~\ref{fig:box}a. Finally, attach the lens (2) and, optionally, the iris (5) to the beamsplitter.


\subsubsection{Computing system}

Connect the other ends of the camera cable (8) to the Raspberry Pi (6). Insert the MicroSD card (7) and connect the power supply (9). If necessary, the LCD monitor screen (10) can also be mounted to the Raspberry Pi for onboard display.

\newpage
\section{Additional Experimental Results}

\subsection{Tolerance to object displacement} Fig.~\ref{fig:sensitivity} quantifies the impact of different object motions on the depth estimation accuracy. We perform this experiment by sequentially capturing $I_1$ and $I_2$ while exerting a known displacement to the object between the capture. We observe the depth estimation algorithm is extremely sensitive to lateral translation and rotations of different axes while being relatively robust to axial translation. Although the study is only conducted using the proposed depth estimation algorithm, it can be translated to other algorithms in the DfDD family due to similarities in solutions. This analysis strongly argues the importance of a snapshot measurement system for DfDD.

\begin{figure}[!h]
    \centering
    \includegraphics[width=\linewidth]{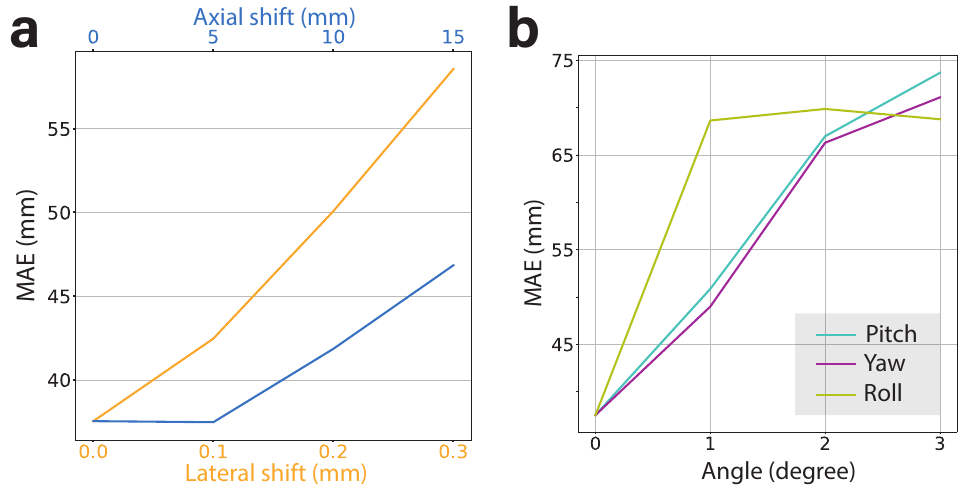}
    \caption{Depth prediction accuracy as a function of relative displacements between $I_1$ and $I-2$ if the two images are captured sequentially. (a) The mean absolute error (MAE) of depth prediction is when the target translates axially or laterally in different amounts. (b) The MAE of depth prediction when the target rotates around different axes. Based on the plots, the MAE increases significantly when a slight displacement between $I_1$ and $I_2$ appears. This result suggests the sensitivity of the proposed method to the relative motion between $I_1$ and $I_2$, as a $0.3$mm lateral shift and a $3^\circ$ rotation doubles the MAE, highlighting the importance of the snapshot capture of $I_1$ and $I_2$. The analysis is based on the real captured data.}
    \label{fig:sensitivity}
\end{figure}

\end{document}